\documentclass[runningheads]{llncs}
\usepackage[T1]{fontenc}
\usepackage{graphicx,verbatim}
\usepackage{booktabs}
\begin{document}
\title{DBMF: A Dual-Branch Multimodal Framework for Out-of-Distribution Detection}
\begin{comment}  %% Removed for anonymized MICCAI submission
\author{First Author\inst{1}\orcidID{0000-1111-2222-3333} \and
Second Author\inst{2,3}\orcidID{1111-2222-3333-4444} \and
Third Author\inst{3}\orcidID{2222--3333-4444-5555}}
%
\authorrunning{F. Author et al.}
\institute{Princeton University, Princeton NJ 08544, USA \and
Springer Heidelberg, Tiergartenstr. 17, 69121 Heidelberg, Germany
\email{lncs@springer.com}\\
\url{http://www.springer.com/gp/computer-science/lncs} \and
ABC Institute, Rupert-Karls-University Heidelberg, Heidelberg, Germany\\
\email{\{abc,lncs\}@uni-heidelberg.de}}

\end{comment}

\author{Jiangbei Yue\inst{1}, Darren Treanor\inst{1}, Venkataraman Subramanian\inst{1},  Sharib Ali\inst{1}\thanks{Corresponding author: S.S.Ali@leeds.ac.uk}}  %% Added for anonymized MICCAI submission
\authorrunning{Jiangbei Yue et al.}
\institute{1 University of Leeds, LS2 9JT, UK}
  
\maketitle              % typeset the header of the contribution
\begin{abstract}
The complex and dynamic real-world clinical environment demands reliable deep learning (DL) systems. Out-of-distribution (OOD) detection plays a critical role in enhancing the reliability and generalizability of DL models when encountering data that deviate from the training distribution, such as unseen disease cases. However, existing OOD detection methods typically rely either on a single visual modality or solely on image-text matching, failing to fully leverage multimodal information. To overcome the challenge, we propose a novel dual-branch multimodal framework by introducing a text-image branch and a vision branch. Our framework fully exploits multimodal representations to identify OOD samples through these two complementary branches. After training, we compute scores from the text-image branch ($S_t$) and vision branch ($S_v$), and integrate them to obtain the final OOD score $S$ that is compared with a threshold for OOD detection. Comprehensive experiments on publicly available endoscopic image datasets demonstrate that our proposed framework is robust across diverse backbones and improves state-of-the-art performance in OOD detection by up to 24.84\%.   

% Authors must provide keywords and are not allowed to remove this Keyword section.

\end{abstract}
\section{Introduction}

% \begin{enumerate}
%     \item The importance of OOD detection. The clarification of OOD detection in medical imaging. (our task) 
%     \item Existing research related to OOD detection in medical imaging
%     \item Challenges 
%     \item How our framework solves the challenges
%     \item contributions
% \end{enumerate}   

\begin{figure}[t]
\centering
\includegraphics[width=\textwidth]{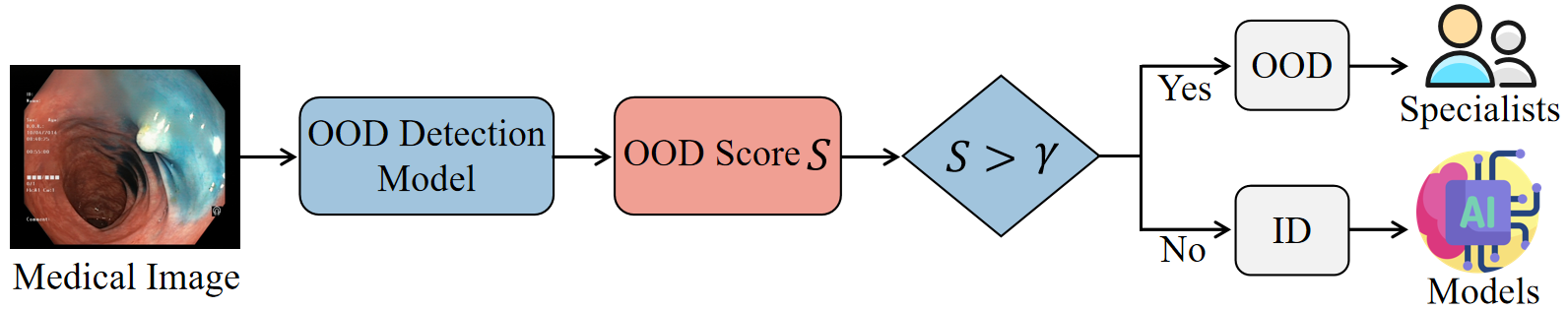}
\caption{The pipeline of the OOD detection. $\gamma$ is a threshold.} \label{fig_intro}
\end{figure}

Deep learning (DL) models are typically trained on limited datasets~\cite{lecun2015deep}, inevitably restricting their coverage of real-world variability. In-distribution (ID) data refer to samples that follow the same distribution as training data and are therefore expected to be handled reliably during deployment. In contrast, out-of-distribution (OOD) data follow a distribution differing from that of training data. Such data often represent unexpected variations that fall outside learned knowledge of models, making reliable prediction challenging~\cite{hendrycks2022scaling}. OOD detection aims to identify the given sample as ID or OOD~\cite{ammar2024neco}, which helps systems avoid making overconfident decisions and instead trigger safeguards such as human review. In this paper, we focus on OOD detection in endoscopic image analysis~\cite{chhetri2025nero}. This task is of critical importance because endoscopic imaging plays a crucial role in medical domains, and numerous related DL models have been developed. However, the reliability of these DL models on OOD data remains questionable. Following previous works~\cite{pokhrel2025out,chhetri2025nero}, we consider normal/healthy and abnormal/unhealthy samples as ID and OOD data, respectively. A common OOD detection pipeline is illustrated in Fig. \ref{fig_intro}. Specifically, a medical image is first fed into an OOD detection model to estimate an OOD score $S$. This score is then compared with a threshold $\gamma$ to determine whether the image is OOD or ID. Finally, ID samples are processed by DL models for automated analysis, while OOD samples are referred to specialists for further evaluation.

Existing research on OOD detection in medical imaging broadly falls into unimodal and multimodal methods. Unimodal methods~\cite{lee2018simple,liu2020energy,huang2021importance} rely solely on a single visual modality and can generally be divided into 3 groups: feature-driven, logit-driven, and gradient-driven approaches. Feature-driven models~\cite{lee2018simple,sun2021react,pokhrel2025out,chhetri2025nero} utilize the distinct characteristics of visual features extracted from ID and OOD samples. For example, NERO~\cite{chhetri2025nero} explores the use of feature-output relevance to enhance the separation between ID and OOD samples. Logit-driven methods~\cite{hendrycks2016baseline,liu2020energy,hendrycks2022scaling} instead focus on prediction logits. Representative approaches calculate the OOD score as the negative maximum logit~\cite{hendrycks2022scaling} or the negative maximum softmax output~\cite{hendrycks2016baseline}. Gradient-driven methods ~\cite{liang2018enhancing,huang2021importance} employ gradient information from neural networks to identify OOD data. GradNorm~\cite{huang2021importance} computes gradients by optimizing the Kullback–Leibler (KL) divergence between softmax predictions and a uniform prior, and uses the negative gradient norm as the OOD score. The rapid advancement of vision-language models (VLMs)~\cite{zhang2024vision}, such as Contrastive Language-Image Pre-training (CLIP)~\cite{radford2021learning}, has stimulated the development of multimodal methods~\cite{ming2024does,ju2025delving}. Ju \textit{et al.}~\cite{ju2025delving} enhanced CLIP with hierarchical prompts and separated ID from OOD samples via image-text alignment. Despite these advances, unimodal methods neglect complementary information from other modalities. Meanwhile, existing multimodal methods primarily rely on image-text matching, without fully exploiting intrinsic visual information beyond cross-modal alignment. 

% Pokhrel \textit{et al.}~\cite{pokhrel2025out} computed OOD scores based on the distances between samples and class centers in the feature space.

To address this challenge, we propose a novel Dual-Branch Multimodal Framework (DBMF), which effectively leverages both textual and visual information. Our framework contains a text-image branch and a vision branch, which are complementary in OOD detection. We train the neural networks in the text-image branch using a new text-separation contrastive loss $L_{TSC}$, which improves the effect of the textual modality. The vision branch is trained by a traditional cross-entropy loss $L_{CE}$~\cite{zhang2018generalized}. After training, we compute scores $S_t$ and $S_v$ from the text-image branch and the vision branch, respectively. The OOD score $S$ is the combination of $S_t$ and $S_v$. Finally, we can utilize $S$ to identify OOD data.         

We evaluate our framework on two public datasets of endoscopic images widely used in OOD detection: Kvasir-v2~\cite{sharma2023deep} and GastroVision~\cite{jha2023gastrovision}. Our framework achieves state-of-the-art (SOTA) performance compared to existing research~\cite{hendrycks2016baseline,liang2018enhancing,lee2018simple,liu2020energy,chan2021entropy,sun2021react,huang2021importance,hendrycks2022scaling,wang2022vim,ammar2024neco,chhetri2025nero}. In particular, our framework shows superior robustness across different network architectures through experiments. Formally, our contributions are summarized as follows. 1) We propose a novel framework, DBMF, which integrates image-text alignment with the vision branch and thereby fully leverages multimodal information. 2) We design a new text-separation contrastive loss $L_{TSC}$ to optimize the image-text branch. 3) We perform extensive evaluations on two public benchmark datasets. The results demonstrate the SOTA and robust performance of the proposed framework.  

% A comprehensive ablation study is provided to validate the complementary advantages of the dual-branch design.
    
% The results show that our framework achieves state-of-the-art performance in OOD detection, maintains strong robustness across diverse backbone architectures, and benefits from the complementary dual-branch architecture.   

% The main contribution is in the novelty of the proposed methodology. 

\section{Methodology}
\begin{figure}[t]
\includegraphics[width=\textwidth]{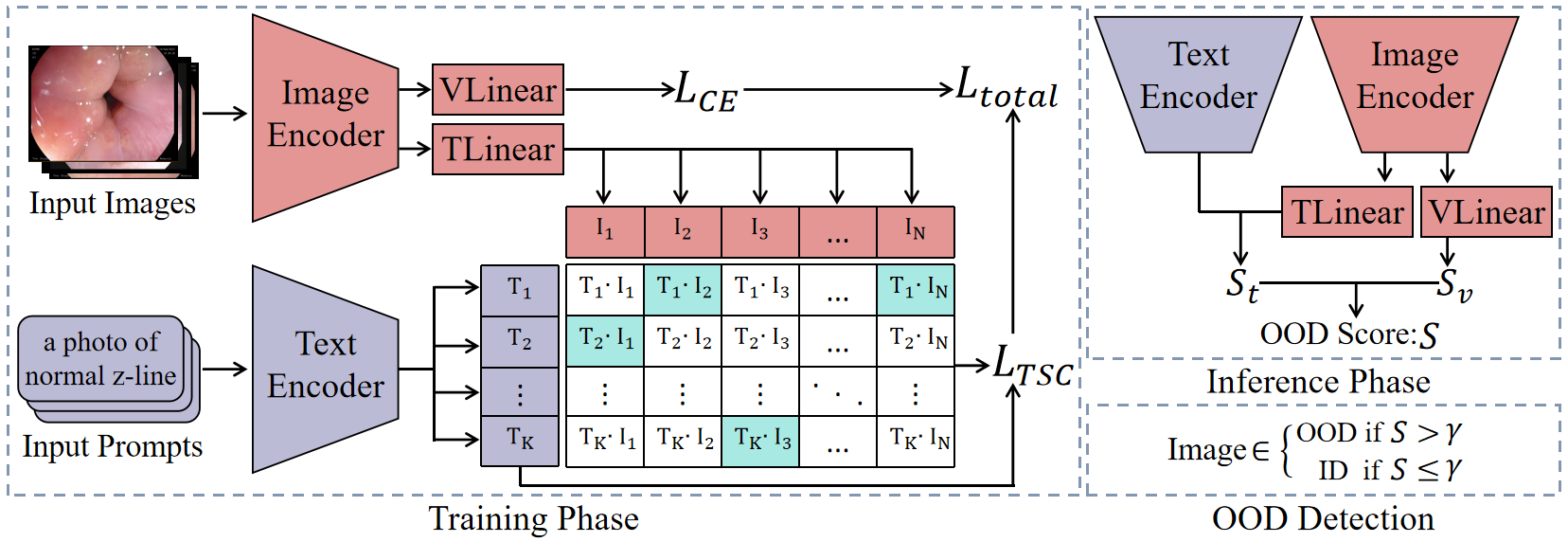}
\caption{The overview of the DBMF. Our framework consists of the training phase, inference phase, and OOD detection. After training two branches, we compute $S_t$ and $S_t$, resulting in the OOD score $S$. The final detection is based on $S$ and a threshold $\gamma$.} \label{fig_overview}
\end{figure}

\textit{Problem Definition}. In the OOD detection problem, we have the training dataset $D_{train}$ from the ID distribution $P_{ID}$, and the testing dataset $D_{test}$ consists of samples from $P_{ID}$ and the OOD distribution $P_{OOD}$. We aim to use $D_{train}$ to train a model that will separate OOD and ID data in $D_{test}$. Following previous research~\cite{hendrycks2016baseline,chan2021entropy,sun2021react,pokhrel2025out}, we train a classification model and calculate OOD scores based on the outputs of the trained model. 

To solve the problem, we propose a new framework DBMF, as shown in Fig. \ref{fig_overview}, which consists of three phases. Given $D_{train} = \{(x_i, y_i)\}_{i=1}^{n_{train}}$, $x_i$ and $y_i$ denote the image and label, respectively. In the training phase, the text-image branch employs the image encoder and text encoder to extract features from input images and prompts, respectively. We use the prompt template of ``\textit{a photo of normal \{class name\}}'', where the class names are from the training data, \textit{i.e.} the distribution $P_{ID}$. Then, the TLinear, which is a linear layer, takes the output of the image encoder as input to generate image features $\{I_i\}_{i=1}^{N}$, where $N$ denotes the number of samples in a batch. The output of the text encoder is the text features $\{T_i\}_{i=1}^{K}$, where $K$ denotes the number of classes in the training dataset. Then, we utilize the text-separation contrastive loss $L_{TSC}$ based on the $\{I_i\}_{i=1}^{N}$ and $\{T_i\}_{i=1}^{K}$ to train two encoders and TLinear. The vision branch places the VLinear, which is also a linear layer, after the image encoder to form a classification model, which is trained by using the cross-entropy loss $L_{CE}$. Therefore, our total loss $L_{total}$ consists of $L_{TSC}$ and $L_{CE}$. After training, we calculate the scores $S_t$ and $S_v$ from the text-image branch and the vision branch, respectively, in the inference phase. The final OOD scores $S$ are obtained by combining $S_t$ with $S_v$. Finally, we conduct OOD detection by calculating OOD scores for images. An image is regarded as OOD if its OOD score exceeds the threshold $\gamma$. Otherwise, we think it is an ID sample.

In our framework, the text-image branch is a CLIP-style model that consists of two encoders projecting images and texts into a shared feature space. The image encoder is commonly implemented using either convolutional neural networks (CNNs)~\cite{he2016deep} or vision transformers~\cite{dosovitskiy2020vit}, while the text encoder is typically a transformer-based language model~\cite{devlin2019bert,radford2021learning}. The TLinear is used to ensure that image and text features have the same dimensionality. We aim to utilize the prototypes of text features to distinguish ID data from OOD data. The introduction of textual information in the text-image branch and the pure visual information in the vision branch are complementary in OOD detection. Specifically, the image encoder with VLinear takes the images as input and yields $\{I_i\}_{i=1}^{N}$, while the prompts are fed into the text encoder to produce $\{T_i\}_{i=1}^{K}$. We introduce a prototype $T_i$ for each class in the training data. Therefore, the batch size $N$ is generally not equal to $K$, which differs from the common CLIP models, where image and text features appear in pairs. As a result, we design a new text-separation contrastive loss $L_{TSC}$ to train the model in the text-image branch. Specifically, each image feature $I_i$ of the sample $x_i$ is aligned with the corresponding text feature $T_{y_i}$ by using the contrastive loss:
\begin{equation}
    \label{eq_lc}
    L_C = -\frac{1}{N}\sum_{i=1}^N \log \frac{\exp(s_{iy_i})}{\sum_{j=1}^N \exp(s_{ij})}, \quad s_{ij}=\frac{I_i^{\top} T_j}{\tau},
\end{equation}
where $\tau$ is a learnable temperature parameter, and all features are normalized to the unit length. To encourage the diversity of text prototypes $\{T_i\}_{i=1}^{K}$, we introduce the text-separation loss: 
\begin{equation}
    L_{TS} =\frac{1}{K^2-K} \sum_{i\neq j} (T_i^{\top}T_j - \eta^{\star})^2, \quad \eta^{\star} = \min\max_{i\neq j} T_i^{\top}T_j,  
\end{equation}
where $\eta^{\star}$ denotes the minimum achievable maximum cosine similarity between different prototypes. Fortunately, $\eta^{\star}$ has the explicit solution $\eta^{\star} = -\frac{1}{K-1}$ \cite{cohn2016packing}. Subsequently, we have the text-separation contrastive loss $L_{TSC} = L_C + \lambda L_{TS}$ in the text-image branch, where $\lambda$ is a balance hyper-parameter. In the vision branch, we train the classification model consisting of the image encoder and VLinear through the cross-entropy loss $L_{CE}$. We generally conduct the training of the vision branch after the training of the text-image branch. Finally, $L_{TSC}$ and $L_{CE}$ together form the total loss $L_{total}$ in our framework. 

After training, we calculate scores $S_t$ and $S_v$ from the text-image and vision branches, respectively. For a test image $x_i$, its feature $I_i$ from the TLinear and the text prototypes $\{T_i\}_{i=1}^{K}$ are utilized to calculate $S_t$:  
\begin{equation}
    S_t =\min_j (-s_{ij}) -  [\sum_{j=1}^K (-s_{ij}) - \min_j (-s_{ij})] = 2 \min_j (-s_{ij}) - \sum_{j=1}^K (-s_{ij}),
\end{equation}
where logits $\{ s_{ij} \}_{j=1}^K$ are from Eq. \ref{eq_lc}. The ID samples generally have low $\min_j (-s_{ij})$ and high $\sum_{j=1}^K (-s_{ij}) - \min_j (-s_{ij})$ because the model is confident in the ID classification, resulting in a low $S_t$. In contrast, we typically have a high $S_t$ for OOD samples. Inspired by~\cite{lee2018simple}, the score $S_v$ is computed based on the Mahalanobis distance, which is effective in OOD detection for the softmax classifier. We use $v$ to denote the visual features from the image encoder in the vision branch. Given the features $\{ v_i \}_{i=1}^{n_{train}}$ extracted from training data, we calculate the mean of each class $\{\mu_k\}_{i=1}^K$ and a shared covariance matrix $\Sigma$ for stability:      
\begin{equation}
    \mu_k = \frac{1}{n_k} \sum_{i=1}^{n_k} v_i^k, \quad \Sigma = \frac{1}{n_{train}} \sum_{k=1}^{K} \sum_{i=1}^{n_k} (v_i^k - \mu_k)(v_i^k - \mu_k)^\top,
\end{equation}
where $n_k$ represents the number of samples within the class $k$ and $n_{train} = \sum_{k=1}^K n_k$. Subsequently, the score $S_v$ for a test image $x_i$ is obtained by:
\begin{equation}
    S_v = \min_k (v(x_i) - \mu_k)^\top \Sigma^{-1}(v(x_i) - \mu_k),
\end{equation}
where $v(x_i)$ denotes the visual feature of $x_i$. We further standardize both $S_t$ and $S_v$, which are transformed into a standard normal distribution. Finally, the OOD score $S$ is determined via $S = S_t + \omega S_v$, where $\omega$ is a hyper-parameter to balance scores from two branches.

% We use  $-s_{ij}$ so that higher scores correspond to OOD samples, consistent with common OOD scoring conventions~\cite{chhetri2025nero}.

% The image encoder is typically implemented using either convolutional neural networks, such as ResNet variants, or Vision Transformers (ViTs), with the latter becoming the dominant choice due to their superior scalability and representation capability under large-scale pretraining. The text encoder is commonly a Transformer-based language model that converts tokenized text prompts into semantic embeddings aligned with image features. During training, both encoders are optimized to maximize the similarity between matched image–text pairs while separating mismatched pairs, enabling effective cross-modal representation learning.

\section{Experiments}
\subsection{Experimental Setup and Implementation Details}
\textbf{Datasets.} Two public endoscopy datasets are used for evaluation: Kvasir-v2~\cite{sharma2023deep} and GastroVision~\cite{jha2023gastrovision}. Kvasir-v2 contains 8,000 images distributed evenly across 8 classes. Following existing research~\cite{chhetri2025nero,pokhrel2025out}, 3 classes of normal anatomical landmarks are treated as ID data, and the remaining 5 classes related to pathological findings and polyp removal as OOD data. ID data are randomly split 8:2, with 80\% used for training and 20\% combined with all OOD samples for testing. GastroVision contains 8,000 images across 27 classes with an imbalanced distribution. Following the previous protocol~\cite{chhetri2025nero,pokhrel2025out}, we treat 3 classes of normal findings and 8 classes of anatomical landmarks as ID data, while 11 classes of pathological findings and 5 classes of therapeutic interventions are considered as OOD data. The construction of the training and testing datasets is identical to that of Kvasir-v2. 

% The images exhibit substantial variation in appearance, illumination, and viewpoint.
% collected from real clinical examinations

\textbf{Metrics.} We employ two evaluation metrics widely used in OOD detection: AUROC~\cite{fawcett2006introduction} and FPR95~\cite{wang2022vim}. AUROC (Area Under the Receiver Operating Characteristic curve) measures the trade-off between the true positive rate and the false positive rate across all possible decision thresholds, indicating the overall model performance. Higher AUROC means better performance. FPR95 denotes the false positive rate (FPR) when the true positive rate is at 95\%, where we typically view ID data as positive samples. Lower FPR95 indicates better performance. In this paper, we report the two metrics in percentage.

% It is the ratio of misclassified OOD samples to all OOD data when most ID samples are correctly detected.

\textbf{Implementation.} In our framework, we typically use CNNs or vision transformers as the image encoder. ResNet18~\cite{he2016deep} and DeiT (Data-efficient Image Transformer)~\cite{touvron2021training} are used as the backbone of the image encoder in the following experiments. The text encoder is generally implemented as a transformer-based language model. We adopt the text transformer from the CLIP~\cite{radford2021learning} as the text encoder for evaluation, which is a 12-layer and 512-wide model with a vocabulary size of 49,408 and 8 attention heads. We typically train the text-image branch before the vision branch. The balance hyper-parameters $\lambda$ and $\omega$ are in the range of $\left[1,1.5\right]$ and $\left[1,3\right]$, respectively.

\subsection{Quantitative Results}
\begin{table}[t]
\centering
\caption{The comparison of experimental results on two datasets across different backbones and metrics. We show the best results in bold and underline the second-best results. Higher AUROC or lower FPR95 indicates better OOD detection performance.}\label{tab_quant}
\begin{tabular}{l|cccccccc}
\toprule
Backbone & \multicolumn{4}{c}{ResNet18} & \multicolumn{4}{c}{DeiT} \\
\midrule

Dataset & \multicolumn{2}{c}{Kvasir-v2} & \multicolumn{2}{c}{GastroVision}
& \multicolumn{2}{c}{Kvasir-v2} & \multicolumn{2}{c}{GastroVision} \\
\midrule
Metric & AUROC & FPR95 & AUROC & FPR95 & AUROC & FPR95 & AUROC & FPR95 \\
\midrule
MSP           & 90.30  & 41.72 & 66.93 & 90.56 & 87.05 & 40.18 & 70.00  & 90.74 \\
ODIN          & \underline{91.77} & 35.44 & 69.79 & 79.27 & 88.41 & 36.40  & 73.37 & 83.68 \\
Mahalanobis   & 84.05 & 54.06 & 65.93 & 89.69 & \underline{94.50} & 21.86 & 75.68 & 81.43 \\
Energy        & 88.85 & 52.36 & 70.31 & 79.79 & 85.77 & 44.02 & 75.35 & 83.68 \\
Entropy       & 90.38 & 41.86 & 67.37 & 87.32 & 87.20  & 39.94 & 70.34 & 90.19 \\
Energy+ReAct  & 86.57 & 53.78 & 61.93 & 83.86 & 83.49 & 46.84 & 73.42 & 83.22 \\
GradNorm      & 85.33 & 54.68 & 62.55 & 90.50  & 71.33 & 57.80  & 54.85 & 88.68 \\
MaxLogit      & 88.90  & 52.38 & 70.08 & 80.44 & 85.77 & 44.02 & 75.11 & 84.20  \\
ViM           & 90.62 & 41.10  & 72.70 & 76.98 & 93.88 & 24.38 & 76.69 & 78.37 \\
NECO          & 89.64 & 47.90 & \underline{79.81} & \underline{71.61} & 88.31 & 37.60 & 76.95 & 81.92 \\
NERO & 90.76 & \underline{28.84} & 75.95 & 74.33 & 92.73 & \underline{18.96} & \underline{82.03} & \underline{76.74} \\
Ours & \textbf{92.11}  & \textbf{25.70}  & \textbf{81.51}  & \textbf{67.60}  & \textbf{94.55}  & \textbf{18.40}  & \textbf{85.84} & \textbf{51.90} \\
\bottomrule
\end{tabular}
\end{table}

For quantitative evaluation, we select 11 SOTA methods as baselines, including MSP~\cite{hendrycks2016baseline}, ODIN~\cite{liang2018enhancing}, Mahalanobis~\cite{lee2018simple}, Energy~\cite{liu2020energy}, Entropy~\cite{chan2021entropy}, ReAct~\cite{sun2021react}, GradNorm~\cite{huang2021importance}, MaxLogit~\cite{hendrycks2022scaling}, ViM~\cite{wang2022vim}, NECO~\cite{ammar2024neco}, and NERO~\cite{chhetri2025nero}. We compare the proposed framework with these baseline methods on Kvasir-v2 and Gastrovision. The experimental results are shown in Table~\ref{tab_quant}, where the best results are shown in bold, and the second-best results are underlined. To ensure a fair comparison, all methods are evaluated using the same image encoder backbone, specifically ResNet18 and DeiT, both pre-trained on Imagenet~\cite{russakovsky2015imagenet}. The experimental results of baselines are cited from~\cite{chhetri2025nero}. Overall, our method achieves SOTA performance across all settings. Notably, the improvements are particularly clear on the more challenging GastroVision dataset, where our method substantially increases AUROC while simultaneously reducing FPR95. Especially, our method improves the AUROC and FPR95 by 3.81\% and 24.84\% with DeiT in GastroVision, respectively. The consistent performance improvement across different backbone architectures demonstrates the robustness of the proposed framework. In summary, these results indicate that our framework provides more reliable separation between ID and OOD data, leading to SOTA OOD detection performance in medical imaging.

\subsection{Qualitative Results}

\begin{figure}[t]
\includegraphics[width=\textwidth]{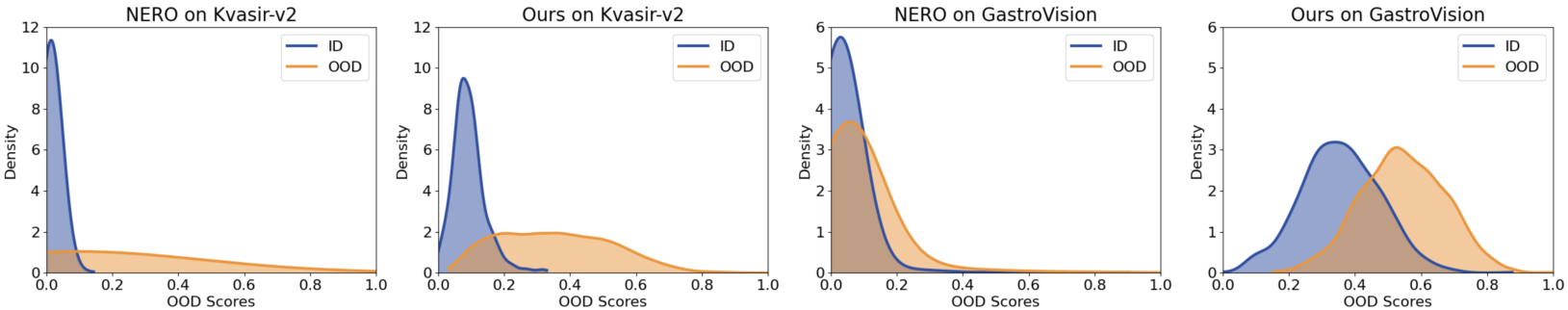}
\caption{Qualitative comparison of OOD score distributions on Kvasir-v2 and GastroVision between NERO and our framework. OOD scores are plotted along the horizontal axis, while the vertical axis shows the corresponding probability density.} \label{fig_quality}
\end{figure}

We also qualitatively compare our framework with the best baseline NERO through visualization of distributions of OOD scores on the testing dataset. We exhibit the results using the backbone of DeiT on Kvasir-v2 and GastroVision in Fig. \ref{fig_quality}, where the OOD score distributions of ID and OOD data are shown in blue and orange, respectively. The horizontal axis represents the OOD scores, and the vertical axis shows the corresponding probability density. For visualization purposes, the OOD scores are linearly rescaled to $\left[0, 1\right]$ via min-max normalization. We estimate the probability density function of the OOD scores through kernel density estimation using Gaussian kernels. 

The two graphs on the left in Fig. \ref{fig_quality} are the OOD score distributions of NERO and our framework on Kvasir-v2. Both of them have low density intersection. However, the scores for ID and OOD data are concentrated near zero for NERO. In contrast, our framework produces a much clearer distributional separation, where scores of ID data remain tightly clustered at low values, while scores of OOD data consistently shifted toward higher values. The right two graphs in Fig. \ref{fig_quality} show the comparison of distributions on the more challenging GastroVision.  For NERO, the concentration of scores near zero becomes more serious with substantial overlap between the two density curves, indicating limited discriminative power and explaining suboptimal OOD detection performance. By contrast, our framework still yields a clear separation of distributions. Although the overlap regions of both methods on GastroVision increase compared with those on Kvasir-v2, our overlap is significantly lower than that of NERO on GastroVision. Overall, the qualitative comparison demonstrates that our framework achieves a more distinct score separation between ID and OOD samples than the best baseline NERO, which aligns with improved quantitative results.

\subsection{Ablation Study}
\begin{table}[t]
\centering
\caption{Ablation study on two datasets. The best results are in bold.}\label{tab_ablation}
\begin{tabular}{l|cccccccc}
\toprule
Backbone & \multicolumn{4}{c}{ResNet18} & \multicolumn{4}{c}{DeiT} \\
\midrule

Dataset & \multicolumn{2}{c}{Kvasir-v2} & \multicolumn{2}{c}{GastroVision}
& \multicolumn{2}{c}{Kvasir-v2} & \multicolumn{2}{c}{GastroVision} \\
\midrule
Metric & AUROC & FPR95 & AUROC & FPR95 & AUROC & FPR95 & AUROC & FPR95 \\
\midrule
Text-image \quad & 88.11 & 42.60  & 78.97  & 77.72  & 92.28  & 34.52  & 82.12  & 63.96  \\
 Vision & 91.50  & 30.40  & 74.31  & 76.27  & 92.09  & 22.86  & 82.51  & 60.51 \\
DBMF & \textbf{92.11}  & \textbf{25.70}  & \textbf{81.51}  & \textbf{67.60}  & \textbf{94.55}  & \textbf{18.40}  & \textbf{85.84} & \textbf{51.90} \\
\bottomrule
\end{tabular}
\end{table}

% better improvement on FPR95

We conduct a comprehensive ablation study to investigate the necessity of the dual-branch architecture in our framework. Specifically, we only retain the text-image branch or the vision branch and evaluate their corresponding performance. The experimental results are compared with those of the full proposed framework, as presented in Table~\ref{tab_ablation}. The results show that using only the text-image branch or only the vision branch yields competitive performance, but both variants consistently underperform against the full framework DBMF. In contrast, DBMF achieves the best AUROC and FPR95 in all evaluation settings, demonstrating that combining two branches provides complementary benefits for OOD detection. The performance gains across both ResNet18 and DeiT backbones indicate the superior robustness of generalization of our DBMF. Overall, these results confirm that each branch contributes useful information, while their integration leads to the best OOD detection performance.

\section{Conclusion}
We propose a novel framework, DBMF, for OOD detection in medical imaging. DBMF consists of a text-image branch and a vision branch, which achieves highly effective multimodal modeling. Comprehensive experiments on two benchmark datasets demonstrate that the proposed framework outperforms existing methods with strong robustness. In the future, we plan to introduce prompt learning~\cite{zhou2022learning} or utilize large language models~\cite{chang2024survey} to generate efficient prompts instead of fixed prompt templates. We also would like to explore the application of pre-trained medical CLIP-style models~\cite{khattak2024unimed} in our framework, which could further enhance the performance in OOD detection.

\bibliographystyle{splncs04}
\bibliography{mybibliography}




% LNCS Homepage, \url{http://www.springer.com/lncs}, last accessed 2023/10/25
% \end{thebibliography}

\end{document}